\documentclass[10pt,twocolumn,letterpaper]{article}

\usepackage{cvpr}      % To produce the REVIEW version
\usepackage{graphicx}
\usepackage{booktabs}

\usepackage{lipsum}
\usepackage{multirow}
\usepackage{colortbl}
\usepackage{amsmath}
\usepackage{caption}
\usepackage{verbatim}  
\usepackage{caption}
\usepackage{makecell}
\usepackage{pifont}
\usepackage{tcolorbox}
\tcbuselibrary{breakable}
\newtcolorbox{promptbox}[1][]{
  breakable,
  colback=blue!3,
  colframe=blue!70,
  boxrule=0.4pt,
  arc=2pt,
  left=4pt,
  right=4pt,
  top=4pt,
  bottom=4pt,
  fontupper=\ttfamily\small,
  title=#1
}

\definecolor{cvprblue}{rgb}{0.21,0.49,0.74}
\usepackage[pagebackref,breaklinks,colorlinks,allcolors=cvprblue]{hyperref}

\def\paperID{*****} % *** Enter the Paper ID here
\def\confName{CVPR}
\def\confYear{2026}

\title{MegaAvatar: Controllable Talking Avatar Generation}

\author{Junyao Gao$^{1,2}$\textsuperscript{*} \quad Sibo Liu$^{1}$\textsuperscript{*} \quad Weidong Zhang$^{1}$ \quad Cairong Zhao$^{2\ddag}$ \quad  Jun Zhang$^{1\ddag}$ \\
	$^{1}$Tencent, $^{2}$Tongji University
	}

\begin{document}
\twocolumn[{%
\renewcommand\twocolumn[1][]{#1}%s
\maketitle
\vspace{-14mm}
\begin{center}
    \centering
    \captionsetup{type=figure}
    \includegraphics[width=\textwidth]{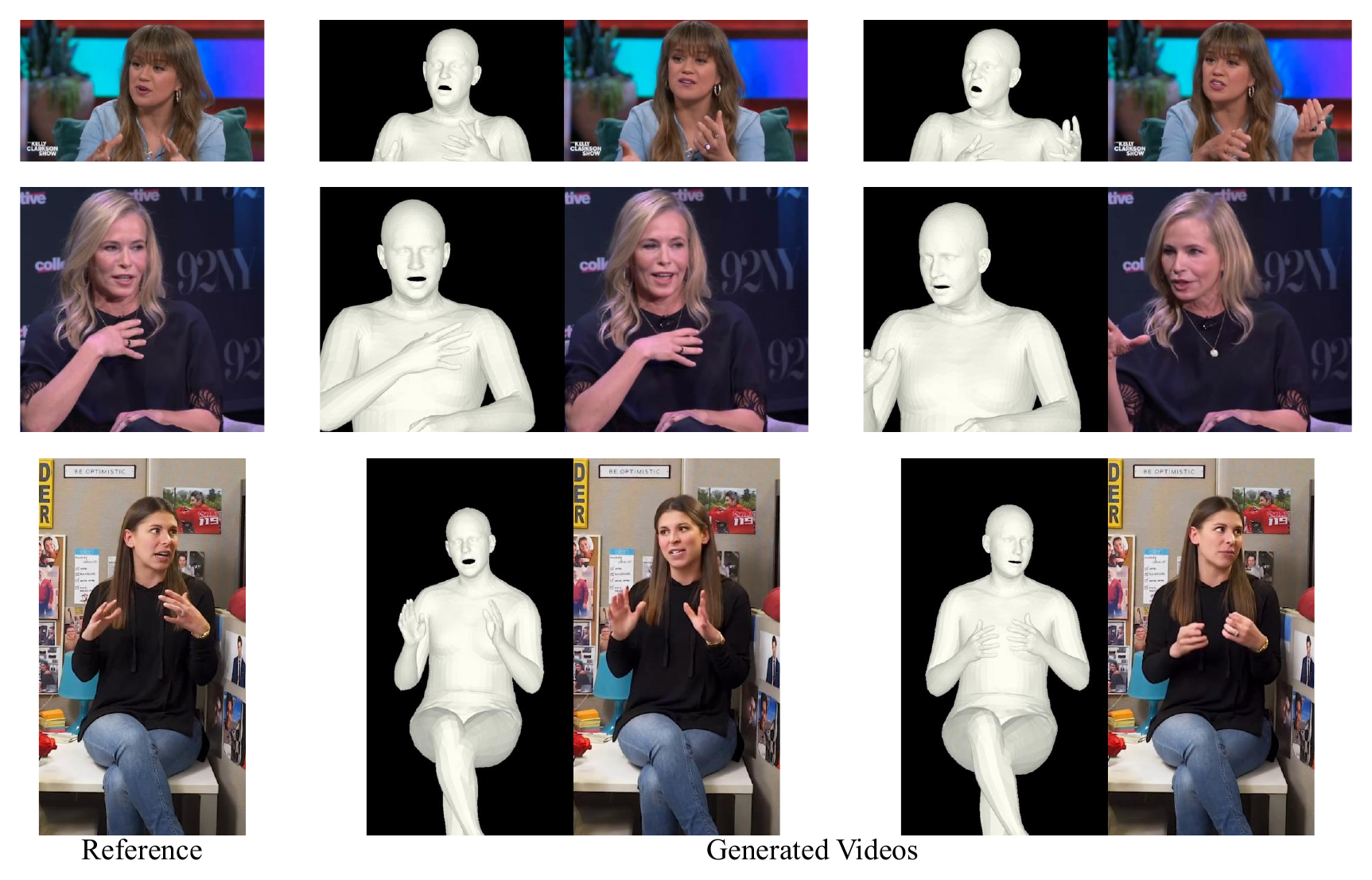}
    \vspace{-8mm}
\captionof{figure}{\textbf{Videos generated by MegaAvatar.} Given a reference image, MegaAvatar generates avatar videos that follow the global motion specified by SMPL-X meshes while producing fine-grained facial expressions synchronized with the input audio.}
\label{fig:teaser}
% \vspace{-0.3cm}
\end{center}}]

\begingroup
\renewcommand\thefootnote{}
\footnote{Work done during Junyao Gao's internship at AIPD, Tencent. \textsuperscript{\ddag}Corresponding authors. *Equal contributions.}
\addtocounter{footnote}{-1}
\endgroup

\begin{abstract}
This report presents \textbf{MegaAvatar}, a controllable talking avatar generation framework built on top of the Wan2.2-TI2V-5B model.
Compared with previous talking-avatar methods that mainly rely on audio or reference-image conditioning, we introduce additional SMPL-X-derived 3D guidance, enabling global control over body pose and head motion.
Specifically, we render the driving SMPL-X sequence into dense mesh frames and encode them with a lightweight 3D convolutional encoder, whose outputs are injected into the latent tokens to provide overall motion control.
Furthermore, we extend Wan2.2-TI2V-5B with additional audio and face cross-attention modules to enable fine-grained expression control and preserve the input identity, respectively.
In addition, we implement an audio-to-SMPL-X model to predict an SMPL-X sequence conditioned on the reference image and input audio, allowing MegaAvatar to support audio-driven inference without user-provided SMPL-X frames.
Experiments show that MegaAvatar achieves high-quality talking avatar generation with controllable body and head motion, speech-synchronized facial expressions, and consistent identity preservation.
MegaAvatar also supports inference with flexible resolutions and video lengths.
Codes, dataset, models will be avaliable in \url{https://github.com/Jeoyal/MegaAvatar}.
\end{abstract}

\begin{figure*}[t]
    \centering
    \includegraphics[width=1.0\linewidth]{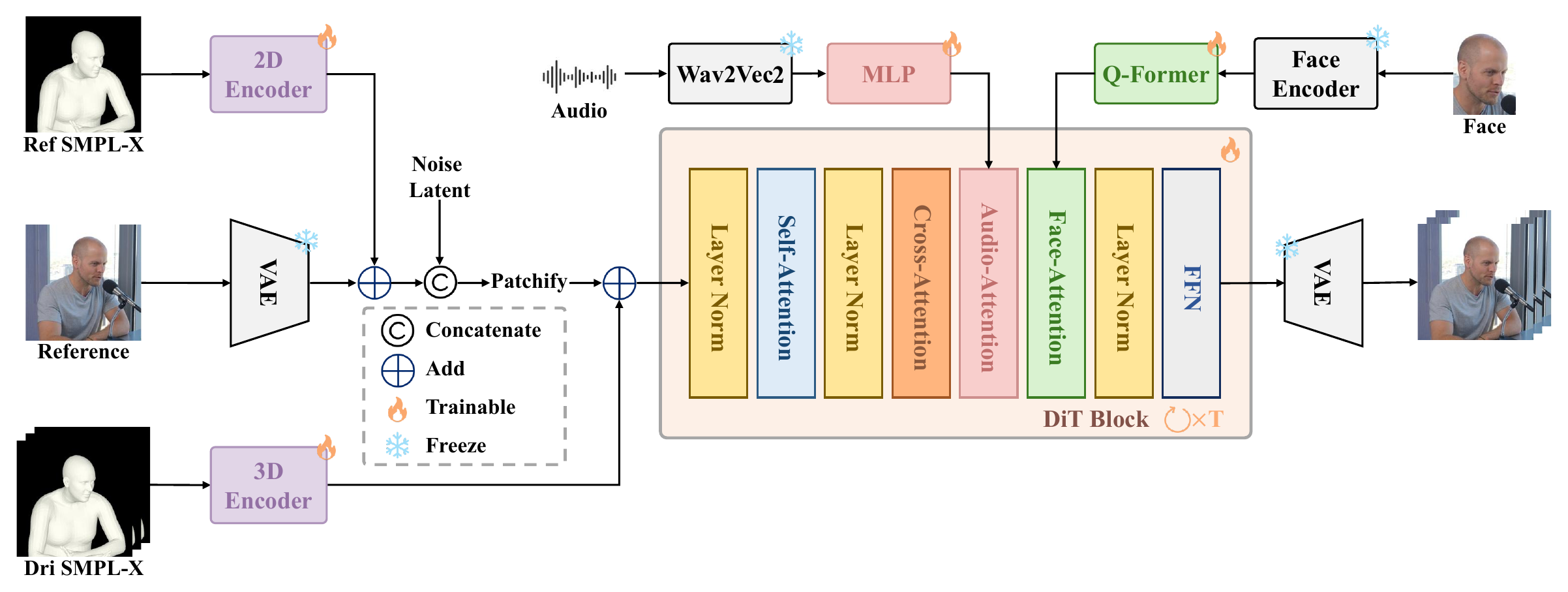}
    \vspace{-8mm}
    \caption{\textbf{Overview of MegaAvatar.} MegaAvatar incorporates SMPL-X, audio, and face conditions into Wan2.2-TI2V-5B for global motion control, fine-grained facial expression control, and identity preservation, respectively.}
    \label{fig:framework}
\end{figure*}

\section{Introduction}
\label{sec:intro}
Talking avatar generation has attracted increasing attention with the rapid development of video generative models. 
This task aims to synthesize a realistic and temporally consistent video of a target person, where the generated avatar should preserve the appearance of a reference image while producing natural body motion, head movement, and speech-related facial expressions. 
Recent diffusion-based video models \cite{blattmann2023stable,wan2025wan,kong2024hunyuanvideo,yang2025cogvideox} have significantly improved the visual quality of generated videos, and large-scale video Diffusion Transformers further provide a strong foundation for high-fidelity human video synthesis.

However, existing talking-avatar methods \cite{team2025klingavatar,jiang2025omnihuman,zeng2026lpm,team2026longcat} are still limited in controllability. 
Most previous methods \cite{cui2025hallo3,wang2025fantasytalking,xu2025hunyuanportrait} mainly rely on audio for mouth movement and speech synchronization, while lacking explicit control over global movements such as body pose and head motion. As a result, their body and head dynamics are difficult to control precisely.

To address this issue, we present MegaAvatar, a controllable talking avatar generation framework built on top of the Wan2.2-TI2V-5B model\footnote{\url{https://huggingface.co/Wan-AI/Wan2.2-TI2V-5B}}.
Compared with previous talking-avatar methods, we introduce additional SMPL-X\cite{SMPL-X:2019}-derived 3D guidance, enabling global control over body pose and head motion. 
During training, MegaAvatar extracts SMPL-X mesh frames from the driving video and encodes them with a lightweight 3D convolutional encoder.
The encoded SMPL-X features are added to the latent tokens, providing global motion control over body pose and head motion.
We also incorporate audio and face conditions through cross-attention modules to enable fine-grained speech-related facial expression control and consistent facial appearance.
In addition, we train an audio-to-SMPL-X model to predict an SMPL-X sequence conditioned on the reference image and input audio.
At inference time, MegaAvatar supports flexible resolutions and variable video lengths.
Moreover, MegaAvatar can operate with only a reference image and an audio clip, or with additional user-provided SMPL-X frames for custom motion control.

Experiments show that MegaAvatar generates high-quality talking avatar videos with globally controllable body and head motion, fine-grained audio-driven facial expressions, and stable identity preservation.
\section{Method}
We present the overall architecture of MegaAvatar in Figure \ref{fig:framework}, which enables controllable talking avatar generation.
MegaAvatar achieves this by incorporating three complementary conditioning signals into the powerful image-conditioned generation model Wan2.2-TI2V-5B:
1) SMPL-X, which provides global motion control over body pose and head motion;
2) Audio, which controls fine-grained speech-related facial expressions and mouth dynamics;
3) Face, which preserves the input identity and stabilizes facial appearance.
In addition, we train an audio-to-SMPL-X model to enable MegaAvatar to support audio-driven inference without user-provided SMPL-X frames.

\noindent\textbf{SMPL-X.}
Given a driving video, we extract the SMPL-X sequence with SMPLer-X \cite{cai2023smpler} and refine the face and hand regions based on EMOCA \cite{danvevcek2022emoca} and HaMeR \cite{pavlakos2024reconstructing}, respectively.
The refined SMPL-X sequence is rendered into dense mesh frames and encoded by a 3D convolutional encoder, whose outputs are added to the patchified latent tokens to provide global motion control over body pose and head motion.
During training, the reference image is randomly sampled from the driving video, VAE-encoded, and prepended as a dedicated reference latent, which serves as the appearance anchor and is excluded from the diffusion target during training.
We also render a reference SMPL-X mesh frame from the reference image and encode it with a 2D convolutional encoder, following \cite{wang2025unianimate}.
The resulting reference feature is added to the reference latent to improve appearance-motion alignment.
Both the 3D and 2D convolutional encoders are trainable.

\noindent\textbf{Audio.}
The audio condition is used to control fine-grained speech-related facial expressions and mouth dynamics.
For each training sample, we extract the corresponding audio clip from the driving video and use a pretrained Wav2Vec2 \cite{baevski2020wav2vec} audio encoder to obtain speech features.
The extracted audio features are then projected into the hidden dimension of Wan2.2-TI2V-5B and injected into the video diffusion transformer through additional audio cross-attention modules.
Instead of using the entire audio clip as a global condition, we adopt frame-level audio conditioning, where each latent frame attends to its temporally aligned audio window.
This design provides more accurate audio-visual alignment and encourages the model to generate mouth movements and facial expressions synchronized with the input speech.
The Wav2Vec2 encoder is kept frozen, while the audio projection and cross-attention modules are trainable.

\noindent\textbf{Face.}
The face condition is used to preserve the input identity and stabilize facial appearance during generation.
Given the reference image, we crop the face region and extract an identity embedding with a pretrained ArcFace encoder.
A Q-Former then converts the identity embedding into identity tokens for face cross-attention, following \cite{wang2025fantasytalking}.
These identity tokens are injected into the video diffusion transformer through additional face cross-attention modules.
In this way, the model receives an explicit identity condition in addition to the reference image, which helps reduce identity drift and maintain consistent facial appearance across frames.
The ArcFace encoder is kept frozen, while the Q-Former and face cross-attention modules are trainable.

\begin{figure}
    \centering
    \includegraphics[width=1.0\linewidth]{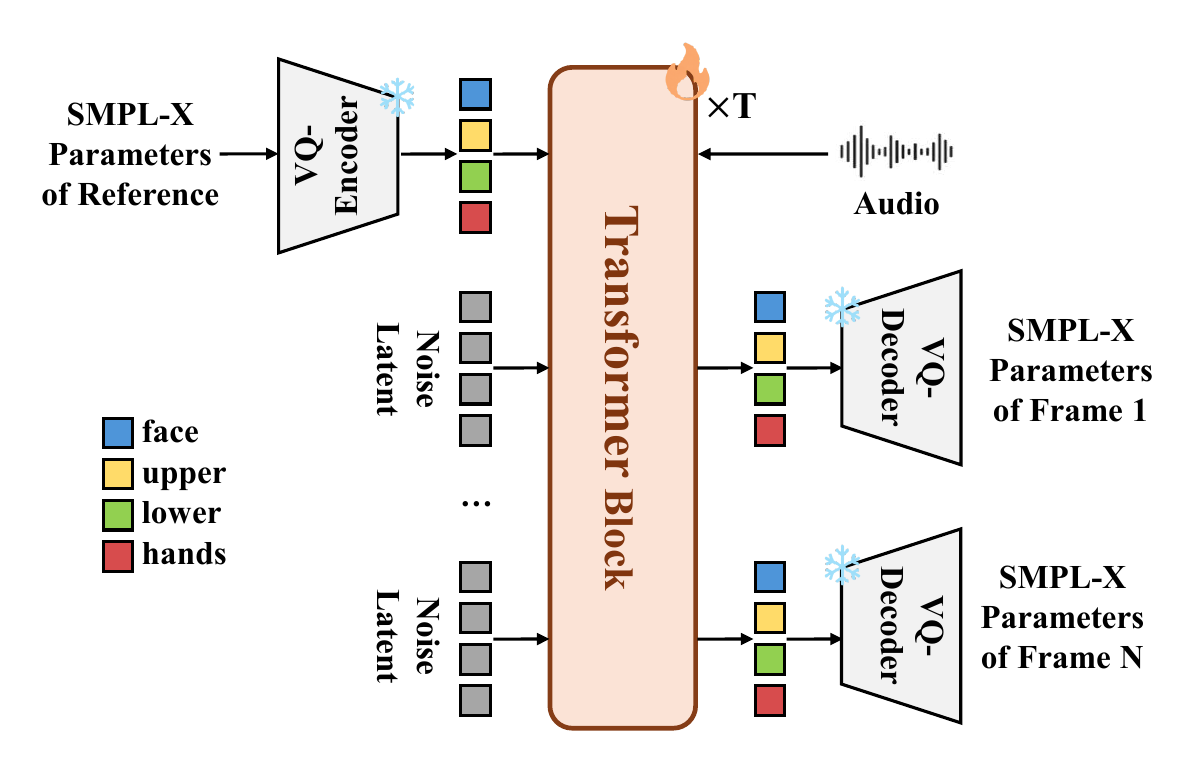}
    \vspace{-5mm}
    \caption{\textbf{Overview of the audio-to-SMPL-X model.} The model predicts SMPL-X motion from the reference SMPL-X and input audio through flow matching in a part-specific VQ latent space.}
    \vspace{-3mm}
    \label{fig:gestruelsm}
\end{figure}

\noindent\textbf{Audio-to-SMPL-X.}
To support audio-driven inference without user-provided SMPL-X frames, we train an audio-to-SMPL-X model to generate the corresponding SMPL-X motion sequence from the reference image and input audio.
We first estimate the SMPL-X parameters of the reference image and convert the SMPL-X motion into a compact motion representation.
Specifically, the upper-body, hands, lower-body/root motion, and face are separately encoded by VQ models, producing part-specific latent representations for full-body motion generation.
We then optimize the audio-to-SMPL-X model with a flow-matching objective in the VQ latent space following \cite{liu2025gesturelsm}.
Given a clean SMPL-X latent and sampled noise, the Transformer denoiser learns to predict the latent velocity conditioned on WavLM \cite{chen2022wavlm} audio features and the encoded SMPL-X representation of the reference image.
At inference time, given only a reference image and the input audio, the audio-to-SMPL-X model predicts the corresponding SMPL-X sequence, which is then rendered into dense mesh frames and used as the motion guidance of MegaAvatar.

\begin{figure}
    \centering
    \includegraphics[width=\linewidth]{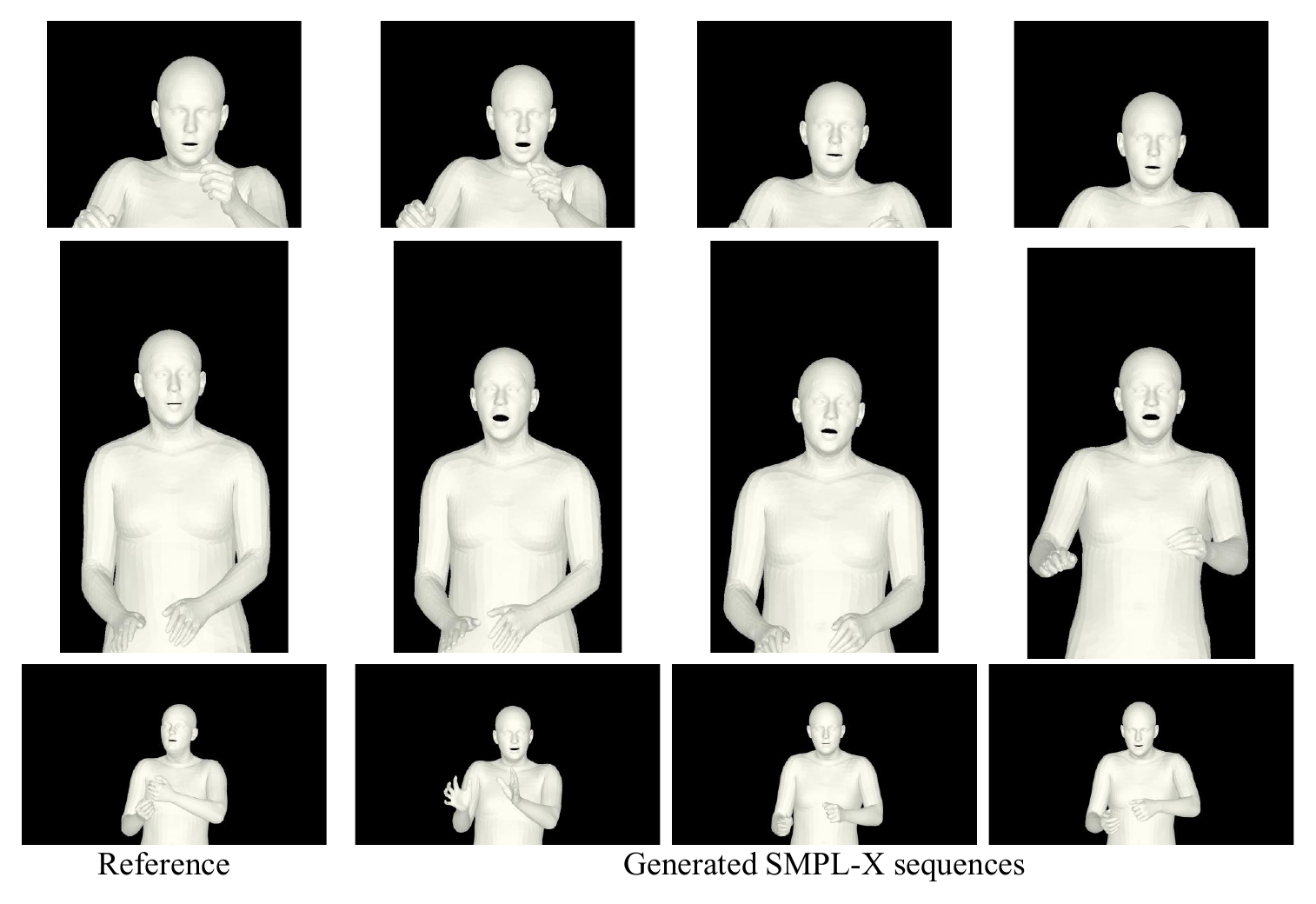}
     \vspace{-8mm}
    \caption{\textbf{Qualitative results of the audio-to-SMPL-X model.} Given a reference image and input audio, our model generates temporally coherent SMPL-X sequences.}
    \vspace{-5mm}
    \label{fig:audio_to_smplx}
\end{figure}

\begin{figure*}
    \centering
    \includegraphics[width=\linewidth]{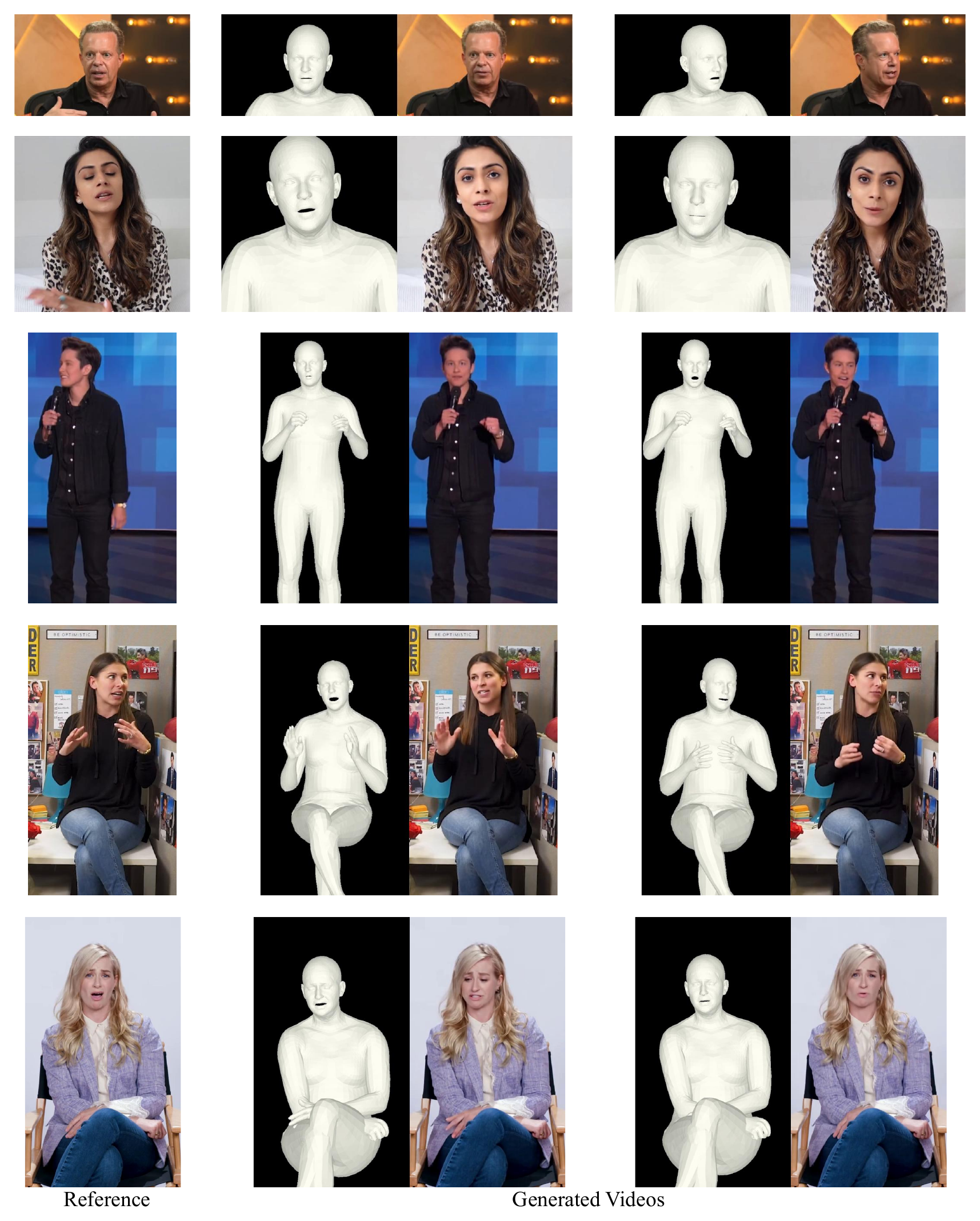}
     \vspace{-8mm}
    \caption{\textbf{Qualitative results of MegaAvatar.} Each row shows a reference image followed by two SMPL-X meshes and their corresponding generated frames.}
    \label{fig:results1}
\end{figure*}

\begin{figure*}
    \centering
    \includegraphics[width=\linewidth]{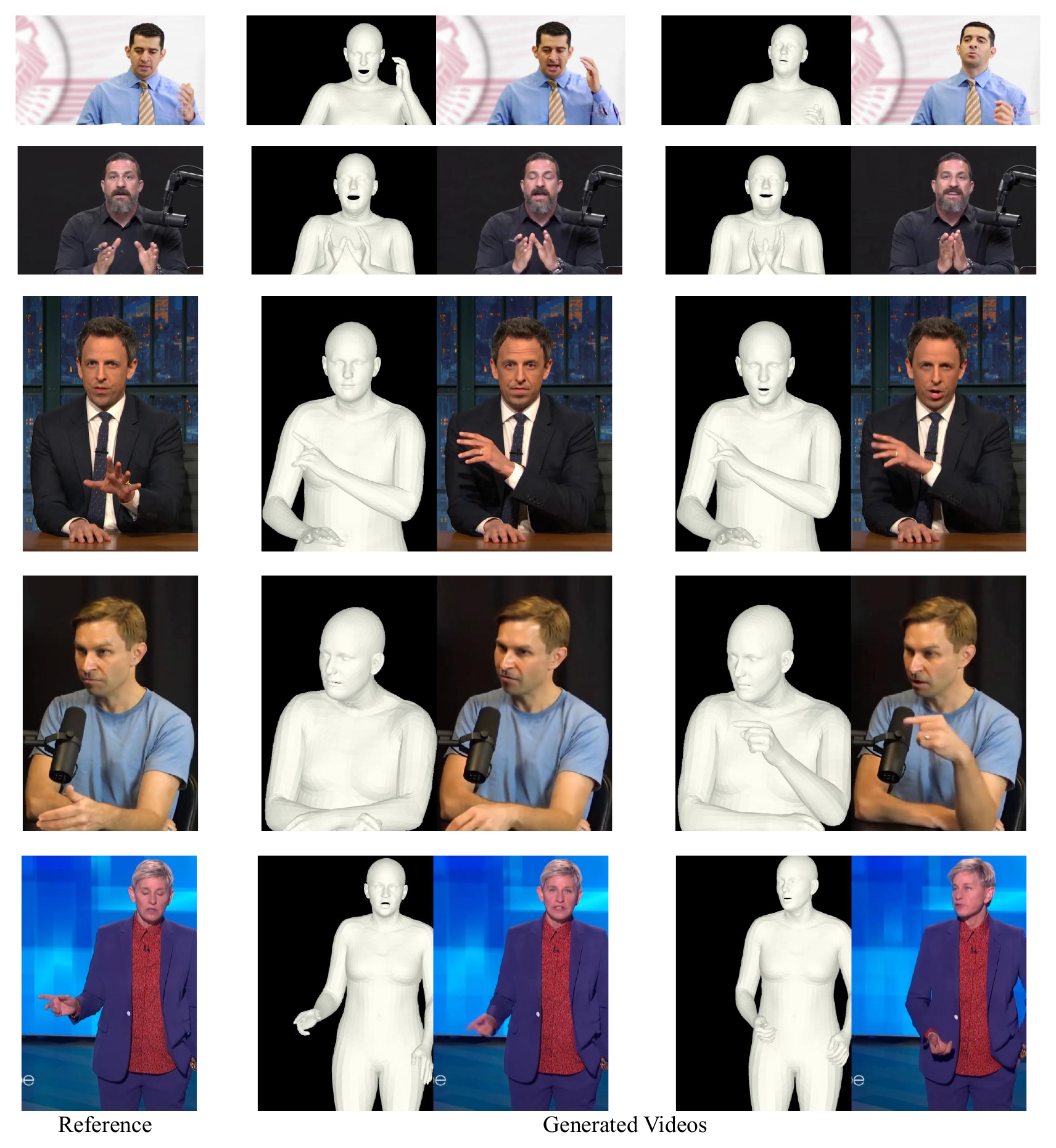}
    \vspace{-8mm}
    \caption{\textbf{Qualitative results of MegaAvatar.} Each row shows a reference image followed by two SMPL-X meshes and their corresponding generated frames.}
    \label{fig:results2}
\end{figure*}

\section{Experiments}
\subsection{Implementation Details}
Following the data construction pipeline of SpeakerVid-5M \cite{zhang2026speakervid}, we collect SpeakerVid-1M, which contains about 1M video clips and 2,000 hours of talking-human videos.
We also process the audio and SMPL-X dense mesh frames for each clip, ensuring that all modalities are temporally aligned at 25 FPS.
We further filter a high-quality subset with single-person videos and paired audio, resulting in SpeakerVid-400K for audio and face training.
We then train MegaAvatar based on Wan2.2-TI2V-5B with the following three-stage training strategy:
In the first stage, we fine-tune the DiT backbone with a rank-128 LoRA \cite{hu2021lora} and the 3D and 2D SMPL-X encoders on SpeakerVid-1M to allow the model to follow dense mesh frames for global motion, including overall body pose and head motion.
Next, we freeze the learned LoRA parameters and the SMPL-X encoders, and train only the audio projection and audio cross-attention modules on SpeakerVid-400K.
To better learn facial expressions and lip dynamics, we increase the flow-matching loss weights on the face and lip regions ($\lambda_{\text{face}}=1.0$, $\lambda_{\text{lip}}=5.0$).
Finally, we keep all previously trained modules frozen and train only the Q-Former and face cross-attention modules on SpeakerVid-400K to improve identity preservation and stabilize facial appearance across frames.
All stages are trained on 16 H20 GPUs with a batch size of 1 per GPU.
The SMPL-X, audio, and face stages are trained for 62K, 96K, and 65K steps, respectively.
We adopt bucketed training with different spatial resolutions and video lengths, allowing MegaAvatar to support flexible-resolution and variable-length inference.

\subsection{Qualitative Evaluation}
We conduct qualitative evaluation on both the intermediate audio-to-SMPL-X results and the final talking avatar videos.
For the audio-to-SMPL-X model, we visualize the predicted SMPL-X sequences as rendered dense mesh frames in Figure \ref{fig:audio_to_smplx}.
The generated SMPL-X frames are temporally smooth and well aligned with the speech rhythm, which enables audio-driven talking avatar generation.

For the final video generation results, MegaAvatar produces high-quality talking avatar videos with clear identity preservation and temporally consistent appearance.
As shown in Figure \ref{fig:results1} and Figure \ref{fig:results2}, the generated videos follow the SMPL-X dense mesh guidance for global body and head motion, while the audio condition controls fine-grained mouth movements and facial expressions.
Moreover, MegaAvatar preserves the input identity and maintains stable facial details throughout the generated video.
These results indicate the effectiveness of the proposed MegaAvatar.

\section{Conclusion}
MegaAvatar advances controllable talking avatar generation by incorporating SMPL-X, audio, and face conditions into Wan2.2-TI2V-5B to enable global control over body pose and head motion, fine-grained speech-related facial expressions, and stable facial appearance, respectively.
Together with the audio-to-SMPL-X model, MegaAvatar supports both audio-driven generation and user-provided SMPL-X control, providing a simple and effective framework for high-quality and controllable talking avatar generation.

{
    \small
    \bibliographystyle{ieeenat_fullname}
    \bibliography{main}

@String(CVPR= {IEEE Conf. Comput. Vis. Pattern Recog.})

@String(ICCV= {Int. Conf. Comput. Vis.})

@String(CVPR  = {CVPR})

@String(ICCV  = {ICCV})

@article{blattmann2023stable,
  title={Stable video diffusion: Scaling latent video diffusion models to large datasets},
  author={Blattmann, Andreas and Dockhorn, Tim and Kulal, Sumith and Mendelevitch, Daniel and Kilian, Maciej and Lorenz, Dominik and Levi, Yam and English, Zion and Voleti, Vikram and Letts, Adam and others},
  journal={arXiv preprint arXiv:2311.15127},
  year={2023}
}

@inproceedings{yang2025cogvideox,
  title={Cogvideox: Text-to-video diffusion models with an expert transformer},
  author={Yang, Zhuoyi and Teng, Jiayan and Zheng, Wendi and Ding, Ming and Huang, Shiyu and Xu, Jiazheng and Yang, Yuanming and Hong, Wenyi and Zhang, Xiaohan and Feng, Guanyu and others},
  booktitle={International Conference on Learning Representations},
  volume={2025},
  pages={83048--83077},
  year={2025}
}

@article{kong2024hunyuanvideo,
  title={Hunyuanvideo: A systematic framework for large video generative models},
  author={Kong, Weijie and Tian, Qi and Zhang, Zijian and Min, Rox and Dai, Zuozhuo and Zhou, Jin and Xiong, Jiangfeng and Li, Xin and Wu, Bo and Zhang, Jianwei and others},
  journal={arXiv preprint arXiv:2412.03603},
  year={2024}
}

@article{wan2025wan,
  title={Wan: Open and advanced large-scale video generative models},
  author={Wan, Team and Wang, Ang and Ai, Baole and Wen, Bin and Mao, Chaojie and Xie, Chen-Wei and Chen, Di and Yu, Feiwu and Zhao, Haiming and Yang, Jianxiao and others},
  journal={arXiv preprint arXiv:2503.20314},
  year={2025}
}

@inproceedings{wang2025fantasytalking,
  title={Fantasytalking: Realistic talking portrait generation via coherent motion synthesis},
  author={Wang, Mengchao and Wang, Qiang and Jiang, Fan and Fan, Yaqi and Zhang, Yunpeng and Qi, Yonggang and Zhao, Kun and Xu, Mu},
  booktitle={Proceedings of the 33rd ACM International Conference on Multimedia},
  pages={9891--9900},
  year={2025}
}

@inproceedings{cui2025hallo3,
  title={Hallo3: Highly dynamic and realistic portrait image animation with video diffusion transformer},
  author={Cui, Jiahao and Li, Hui and Zhan, Yun and Shang, Hanlin and Cheng, Kaihui and Ma, Yuqi and Mu, Shan and Zhou, Hang and Wang, Jingdong and Zhu, Siyu},
  booktitle={2025 IEEE/CVF Conference on Computer Vision and Pattern Recognition (CVPR)},
  pages={21086--21095},
  year={2025},
  organization={IEEE}
}

@inproceedings{xu2025hunyuanportrait,
  title={Hunyuanportrait: Implicit condition control for enhanced portrait animation},
  author={Xu, Zunnan and Yu, Zhentao and Zhou, Zixiang and Zhou, Jun and Jin, Xiaoyu and Hong, Fa-Ting and Ji, Xiaozhong and Zhu, Junwei and Cai, Chengfei and Tang, Shiyu and others},
  booktitle={2025 IEEE/CVF Conference on Computer Vision and Pattern Recognition (CVPR)},
  pages={15909--15919},
  year={2025},
  organization={IEEE}
}

@article{team2025klingavatar,
  title={Klingavatar 2.0 technical report},
  author={Team, Kling and Chen, Jialu and Ding, Yikang and Fang, Zhixue and Gai, Kun and Gao, Yuan and He, Kang and Hua, Jingyun and Jiang, Boyuan and Lao, Mingming and others},
  journal={arXiv preprint arXiv:2512.13313},
  year={2025}
}

@article{jiang2025omnihuman,
  title={Omnihuman-1.5: Instilling an active mind in avatars via cognitive simulation},
  author={Jiang, Jianwen and Zeng, Weihong and Zheng, Zerong and Yang, Jiaqi and Liang, Chao and Liao, Wang and Liang, Han and Zhang, Yuan and Gao, Mingyuan},
  journal={arXiv preprint arXiv:2508.19209},
  year={2025}
}

@article{zeng2026lpm,
  title={Lpm 1.0: Video-based character performance model},
  author={Zeng, Ailing and Yang, Casper and Ge, Chauncey and Zhang, Eddie and Xu, Garvey and Lin, Gavin and Gu, Gilbert and Pi, Jeremy and Li, Leo and Shi, Mingyi and others},
  journal={arXiv preprint arXiv:2604.07823},
  year={2026}
}

@article{team2026longcat,
  title={LongCat-Video-Avatar 1.5 Technical Report},
  author={Team, Meituan LongCat and Cai, Xunliang and Cheng, Meng and Gao, Feng and Kong, Zhe and Li, Jiamu and Li, Le and Li, Weiheng and Liu, Hongyu and Tan, Shuai and others},
  journal={arXiv preprint arXiv:2605.26486},
  year={2026}
}

@inproceedings{SMPL-X:2019,
    title = {Expressive Body Capture: 3D Hands, Face, and Body from a Single Image},
    author = {Pavlakos, Georgios and Choutas, Vasileios and Ghorbani, Nima and Bolkart, Timo and Osman, Ahmed A. A. and Tzionas, Dimitrios and Black, Michael J.},
    booktitle = {Proceedings IEEE Conf. on Computer Vision and Pattern Recognition (CVPR)},
    year = {2019}
}

@article{cai2023smpler,
  title={Smpler-x: Scaling up expressive human pose and shape estimation},
  author={Cai, Zhongang and Yin, Wanqi and Zeng, Ailing and Wei, Chen and Sun, Qingping and Yanjun, Wang and Pang, Hui En and Mei, Haiyi and Zhang, Mingyuan and Zhang, Lei and others},
  journal={Advances in Neural Information Processing Systems},
  volume={36},
  pages={11454--11468},
  year={2023}
}

@inproceedings{danvevcek2022emoca,
  title={Emoca: Emotion driven monocular face capture and animation},
  author={Dan{\v{e}}{\v{c}}ek, Radek and Black, Michael J and Bolkart, Timo},
  booktitle={Proceedings of the IEEE/CVF conference on computer vision and pattern recognition},
  pages={20311--20322},
  year={2022}
}

@inproceedings{pavlakos2024reconstructing,
    title={Reconstructing Hands in 3{D} with Transformers},
    author={Pavlakos, Georgios and Shan, Dandan and Radosavovic, Ilija and Kanazawa, Angjoo and Fouhey, David and Malik, Jitendra},
    booktitle={CVPR},
    year={2024}
}

@article{baevski2020wav2vec,
  title={wav2vec 2.0: A framework for self-supervised learning of speech representations},
  author={Baevski, Alexei and Zhou, Yuhao and Mohamed, Abdelrahman and Auli, Michael},
  journal={Advances in neural information processing systems},
  volume={33},
  pages={12449--12460},
  year={2020}
}

@inproceedings{liu2025gesturelsm,
  title={Gesturelsm: Latent shortcut based co-speech gesture generation with spatial-temporal modeling},
  author={Liu, Pinxin and Song, Luchuan and Huang, Junhua and Liu, Haiyang and Xu, Chenliang},
  booktitle={2025 IEEE/CVF International Conference on Computer Vision (ICCV)},
  pages={10929--10939},
  year={2025},
  organization={IEEE}
}

@article{chen2022wavlm,
  title={Wavlm: Large-scale self-supervised pre-training for full stack speech processing},
  author={Chen, Sanyuan and Wang, Chengyi and Chen, Zhengyang and Wu, Yu and Liu, Shujie and Chen, Zhuo and Li, Jinyu and Kanda, Naoyuki and Yoshioka, Takuya and Xiao, Xiong and others},
  journal={IEEE Journal of Selected Topics in Signal Processing},
  volume={16},
  number={6},
  pages={1505--1518},
  year={2022},
  publisher={IEEE}
}

@article{wang2025unianimate,
  title={Unianimate: Taming unified video diffusion models for consistent human image animation},
  author={Wang, Xiang and Zhang, Shiwei and Gao, Changxin and Wang, Jiayu and Zhou, Xiaoqiang and Zhang, Yingya and Yan, Luxin and Sang, Nong},
  journal={Science China Information Sciences},
  volume={68},
  number={10},
  pages={200103},
  year={2025},
  publisher={Springer}
}

@article{hu2021lora,
  title={Lora: Low-rank adaptation of large language models},
  author={Hu, Edward J and Shen, Yelong and Wallis, Phillip and Allen-Zhu, Zeyuan and Li, Yuanzhi and Wang, Shean and Wang, Lu and Chen, Weizhu},
  journal={arXiv preprint arXiv:2106.09685},
  year={2021}
}

@inproceedings{zhang2026speakervid,
  title={Speakervid-5m: A large-scale high-quality dataset for audio-visual dyadic interactive human generation},
  author={Zhang, Youliang and Li, Zhaoyang and Wang, Duomin and Zhou, Deyu and Yin, Zixin and Dai, Xili and Yu, Gang and Li, Xiu and others},
  booktitle={International Conference on Learning Representations},
  volume={2026},
  pages={117896--117926},
  year={2026}
}
}


\end{document}